# A Supervised Embedding and Clustering Anomaly Detection method for classification of Mobile Network Faults

R. Mosayebi, H. Kia, A. Kianpour Raki

*Abstract*— The paper introduces Supervised Embedding and Clustering Anomaly Detection (SEMC-AD), a method designed to efficiently identify faulty alarm logs in a mobile network and alleviate the challenges of manual monitoring caused by the growing volume of alarm logs. SEMC-AD employs a supervised embedding approach based on deep neural networks, utilizing historical alarm logs and their labels to extract numerical representations for each log, effectively addressing the issue of imbalanced classification due to a small proportion of anomalies in the dataset without employing one-hot encoding. The robustness of the embedding is evaluated by plotting the two most significant principle components of the embedded alarm logs, revealing that anomalies form distinct clusters with similar embeddings. Multivariate normal Gaussian clustering is then applied to these components, identifying clusters with a high ratio of anomalies to normal alarms (above 90%) and labeling them as the anomaly group. To classify new alarm logs, we check if their embedded vectors' two most significant principle components fall within the anomaly-labeled clusters. If so, the log is classified as an anomaly. Performance evaluation demonstrates that SEMC-AD outperforms conventional random forest and gradient boosting methods without embedding. SEMC-AD achieves 99% anomaly detection, whereas random forest and XGBoost only detect 86% and 81% of anomalies, respectively. While supervised classification methods may excel in labeled datasets, the results demonstrate that SEMC-AD is more efficient in classifying anomalies in datasets with numerous categorical features, significantly enhancing anomaly detection, reducing operator burden, and improving network maintenance.

## I. Introduction

In recent decades, with the dramatic growth of the Internet and mobile phone users, telecommunication networks have also expanded so that providers cover more areas and locations. To adapt to this growth, service providers need extensive monitoring within the network to be able to maintain the quality of services and detect faults and problems as soon as possible. This process takes place in some offices called network operations centers (NOC) which is generally responsible for monitoring and managing the network [1]. The operators of the NOC face a crowded dashboard, with an ongoing stream of different alarms, with a small portion leading to a failure or fault. As a result, the operator needs to rummage through the alarms to be able to find the ones that need the transformation to tickets. Due to the process being prone to human mistakes, it is conceivable that some alarms get missed or overlooked by the operator. Furthermore, having a diligent process where each alert is scrupulously reviewed takes a large amount of time and degrades automation. The emergence of IoT devices and, more broadly, 5G connectivity will lead to ever-larger numbers of alarms, which in turn will aggravate the situation mentioned above. Therefore, the telco operator will need a mechanism, or more specifically, an AI capability, to assist with fault generation to eliminate exponential workload on the operators. Besides reducing the response time, reducing the cost, and increasing the efficiency of OSS, the AI agent can be more applicable to the emergence of 5G technology.

By increasing the rate of data transfer and the spread of IoT devices, the rate of alarm occurrence will also increase, in which the traditional methods do not scale up for efficient alarm and fault detection. A fully trained AI agent can classify the alarms more effective in seconds with high identification and resolution.

With the emergence of artificial intelligence and machine learning, various methods are applied to automate the process of anomaly detection and problem ticket creation [2]. The autonomous network assurance reduces the operator work load and the time for problem resolution and network maintenance which consequently increase the customer satisfaction and prevent from customer churn [3].

There are various papers and researches for anomaly detection in different domains of telecommunication. For example, in [4] the author has used the online analytical processing method for real network anomaly detection. In [5] the authors elaborated a combination of support vector machine and LSTM networks to a set of long term observation of periodically collected Key Performance Indicators from a Cellular/Wireless network. In [6], Two main algorithms, ensemble-based Isolation Forest and auto encoder neural network, are employed and applied in order to detect anomalous patterns in key performance indicators of long term evolution (LTE) networks describing the package losses, delays, transmission success rates, etc.. In [7], the author focus is on detecting the abnormalities in the telecommunication domain using the Call Detail Records (CDR). The anomalies are identified using the clustering techniques such as k-means clustering, hierarchical clustering and PAM clustering.

R. Mosayebi is with the School of Electrical and Computer Engineering, College of Engineering, University of Tehran, Tehran, Iran (e-mail: mosayebi@ut.ac.ir) and Head of Artificial Intelligence team, Clarity Global Co., Sydney, Australia. (r.mosayebi@clarityint.com)
H. Kia is with the School of Electrical and Computer Engineering, K.N.Toosi University of Technology, Tehran, Iran (e-mail: kia.hanif@email.kntu.ac.ir) and member of Artificial Intelligence team, Clarity Global Co., Sydney, Australia. (h.kia@clarityint.com)
A. Kianpour Raki is with Artificial Intelligence team, Clarity Global Co., Sydney, Australia. (a.kianpour@clarityint.com)
S. M. Zoee is a member of TMForum and is with Clarity Global Co., Sydney, Australia. (mzoee@clarityint.com)

In this method, we used a deep learning neural network for embedding the alarm stream using the historical alarm dataset with labels of anomaly and normal. Then, the two most principle components of the embedded alarm logs are used in a clustering approach to classify the new alarm stream in two groups of anomaly and normal. Our contribution is summarized as follow:

1. Using the machine learning techniques for anomaly detection of alarm streams, we have increased the network efficiency by reducing the time for anomaly detection which result in faster problem resolution.

2. The proposed methodology has higher performance with respect to ordinary random forest and XGboost in terms of detection of the percentage of anomalies in the alarm logs.

This paper is organized as below. In section2, the features of data and the preprocessing steps are described. In section 3, the anomaly detection problem is defined and the proposed method is explained. In section 4, the result and comparison are illustrated. The paper is then concluded in section 5.

## II. MATERIALS & METHODS

### A. Data & preprocessing

For the learning process, the historical alarm data of the relevant faults from *August 8, 2020 to January 20, 2021* is used. This data is collected from three tables in Clarity database with different structures. After unifying the data structure and turning it into a standard tabular dataset, we had 64 attributes, including reporting and clear time of alarms, location information, NMS information, alarm types, and severity. As each vendor sends the same alarm types with their own standard names, it is necessary to use a ruled-based mapping method to map alarm types with the same concepts into a single category, in the preprocessing steps. **Error! Reference source not found.** shows the main steps of data preprocessing phase.

Different time-based information such as hour, day, weekday, month, season and year are extracted from the reporting times of the alarms, and are used as separate features. For severity attribute, a number is assigned to each entity in order to keep the intensity level. Label encoder was applied on rest of the categorical features.

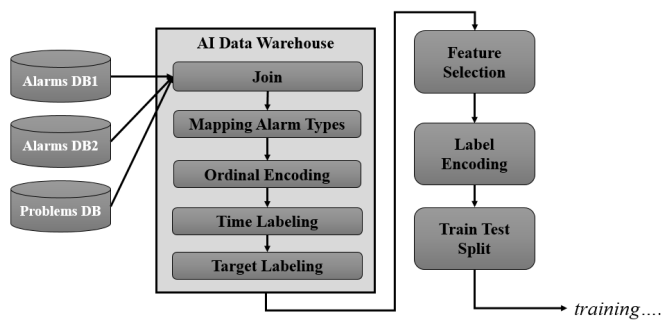

*Fig. 1 Fault Detection preprocessing workflow*

It is also necessary to add a special entity for each feature for handling the values that do not exist in the training data, but may exist in the deployment mode. As the result of this, the label encoder can also recognize the new values in each feature. Then, by statistical tests such as Chi-Square-Test [8] and Theil's U entropy correlation [9], the most important features are selected. These features had the highest correlation with the target feature. The 10 final selected attributes are listed in TABLE 1.

### B. Random Forest & XGBoost

Random forest (RF) [10] [11] and XGBoost [12] are among the tree-based methods which have proper performance when facing with unbalanced structural datasets. RF is an ensemble classifier which means it uses various models as based classifiers (Decision tree), then new classifier is derived, which had better performance in comparison with any constituent classifier. XGBoost (stands for Extreme Gradient Boosting) is an optimized distributed gradient boosting techniques which provides a parallel tree boosting (also known as GBDT, GBM) to solve many problems in a fast and accurate way. The gradient boosting model is a linear combination of some weak models that are periodically constructed to create a strong and accurate final model.

Performance compatible with categorical attributes, overcoming the problem of overfitting, high speed and the ability to work in unbalanced environments are the main reasons that led us to use these two tree-based methods.

*TABLE 1*

*Summary of the selected alarm attributes*

| Attribute | Range |
|---|---|
| Severity | [1,6] |
| Alarm Type | [1,114] |
| Site Code | [1,441] |
| City | [1,111] |
| Domain | [1,9] |
| Segment Name | [1,405] |
| Management System | [1,12] |
| Port Type | [1,12] |
| Equipment Type | [1,18] |
| Hour | [1,24] |

### C. Categorical Feature Embedding

Working with categorical features has always been one of the most challenging parts of machine learning and makes it impossible to perform many calculations. Since in many telecommunication and OSS data (such as our case), we only deal with categorical features with a wide range of variables for each, the mentioned complications become more apparent. Therefore, often the tree-based models; such as Decision trees and Random Forests can be used because of their proper performance to learn the patterns in the structural datasets with categorical features. There are couple of main approaches to deal with categorical data for representing them as vectors: One Hot Encoding and Feature Embedding [13].

In this use case, the one hot encoding is almost impractical because of the high number of unique values for each attribute and hardware limitations. Therefore, feature embedding is considered as a candidate solution. The feature embedding is based on the process of training a Neural Network for classifying the alarms into normal and anomaly groups. The label dataset is used for training this neural network. Several embedding layers are separately applied on the features of the dataset, before the fully dense layers, using the Keras framework [14]. The output of the embedding layers are then concatenated and passed to Fully Connected Neural Networks. During the training of the whole model, the embedding weights are also trained. We are then able to detach the embedding layer from the network and use it as a separate numerical representation of the alarm data (Fig. 2). Therefore, a new representation of our features are obtained in an 1185-dimensional space.

This allows us to have a more compact and informative input when compared to a single One-Hot-Encoding approach. By adopting entity embedding, in addition to solving the problem of the huge number of possible values for each feature, we also are able to mitigate another major problem: It does not need to have a domain expert, once we're capable to train a Neural Network that can efficiently learn patterns and relationships between the values of a same categorical feature. This leads to avoid the feature engineering step (such as manually giving weights to each).

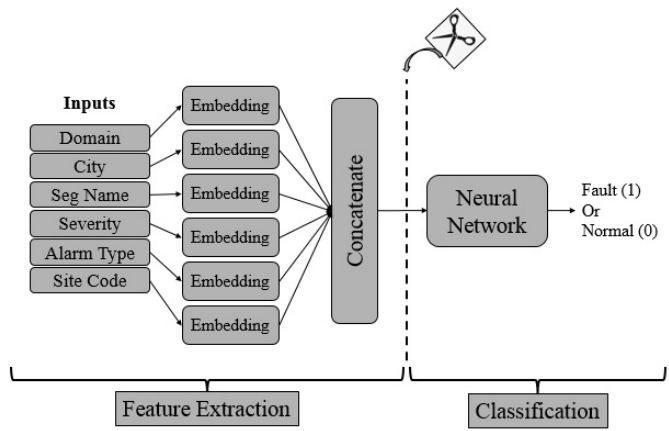

*Fig. 2 Embedding Categorical Features*

An Embedding layer is pretty much a Neural Network layer that groups each values of the categorical features into an N-dimensional space. This spatial representation allows us to obtain the intrinsic properties of each categorical value, extracting the relation between these values while having a compact representation.

*D. Principal Component Analysis*

Principal component analysis (PCA) is simply a way to extract important variables (in the form of components) from a large set of variables in a dataset [15]. A principal component is a normalized linear combination of the main predictions in the dataset. PCA actually extracts a low-dimensional set of features from a high-dimensional set to keep more information with fewer variables. In this way, data visualization also becomes more meaningful. PCA method is more useful when dealing with three or more dimensions. It is always applied on the covariance or correlation matrix, therefore, the data must be numerical and standardized

The PCA algorithm is firstly applied on the embedded dataset to remove noise and artifacts and extract the most significant components which contain the most variance in the dataset. Fig. 3 shows the first 50 extracted variances of features. We used the first two components which have the highest amount of variance for further processing.

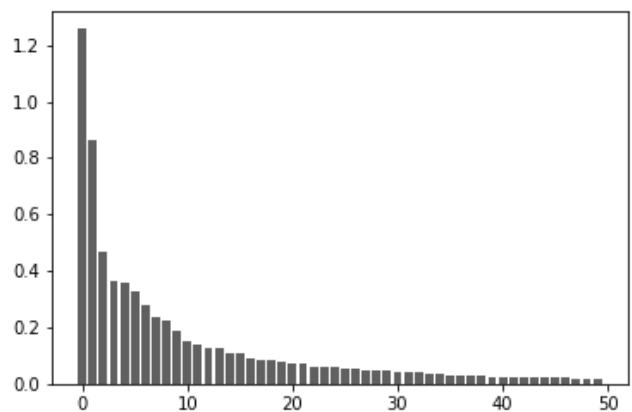

*Fig. 3 Principal Components Variances*

*E. Proposed Method*

In this paper, we propose a Supervised Embedding and Clustering Anomaly Detection (SEMC-AD) method to identify the faulty alarm logs in a fraction of a second, reduce the operator workload and enhance the network maintenance by reducing the time for problem resolution.

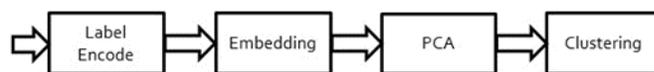

*Fig. 4 Illustration diagram of the SEMC-AD method*

As shown in the Fig. 4 The SEMC-AD method works as follows:
1. An embedding representation of alarm data is obtained with the proposed approach in part C.
2. PCA is applied on the embedded alarm data. The number of significant principle components *n* is selected such that the *n* principle components contain the 90% variance of data.
3. The scatter plot of the principle components are plotted to determine the number of necessary clusters.
4. A proper clustering approach is applied on the principle components from step 2. The clusters in which the ratio of the number of anomalies to normal alarms are higher than 90 percent, are labeled as anomaly group.

5. The new alarm log will be classified as an anomaly if the most significant principle components of its embedded vector are classified in clusters with the label of anomaly.

### III. RESULTS & DISCUSSIONS

*A. Apply baseline methods*

The main goal of the project is to learn an AI-based model which could accurately distinguish faults from a huge number of normal alarms. Due to the severe imbalance distribution of data, it is necessary to evaluate the performance of the model in each class separately and by two criteria, precision and recall.

Since network administrators need to maintain the desired quality of communication networks and services as fast as possible, they must be informed from all faults immediately to take the prompt and necessary actions for mitigating them. As a result, the performance of the AI agent must be very sensitive to the faults, meaning that the final model should have high percentage of recall for anomaly class.

**Error! Reference source not found.** and **Error! Reference source not found.** shows the precision-recall curve of Random Forest and XGboost, respectively. As it can be seen in the diagrams, there is a trade-off between the values of recall and precision. In the other words, increasing the value of Recall results in the reduction of the Precision. Therefore, in order for the model to be capable of detecting a high percentage of faults, some misclassification of normal alarms must be accepted. As a result, model thresholds have been adjusted to reach 60% of precision in detecting fault class and the recall value of the fault class was compared for different methods. TABLE 2 shows performance of these methods.

As the results show, with a constant precision of about 60%, the recall value for random forest and XGBoost are 86% and 81%, respectively. In the following, we present a clustering-based method that has significantly increased the value of recall.

*B. Supervised Embedding and Clustering Anomaly Detection (SEMC-AD)*

Clustering analysis is an unsupervised learning method that separates the data points into several specific bunches or groups, such that the data points in the same groups have similar properties and data points in different groups have different properties in some sense. In the proposed method, we used two methods of clustering, K-Means [16] and Gaussian Mixture Model (GMM) [17]. K-means works based on the Euclidean distance between the data points while GMM is a density-based method, looking for Gaussian distributions among feature space (Fig. 7).

Since our labels were overlapped in many parts of the feature space, a proper separation of the labels is not feasible by distance-based methods such as K-means. On the other hand, GMM reveals the density of the labels in different parts of the feature space and provides a more appropriate and acceptable separation. Therefore, we used this method for clustering.

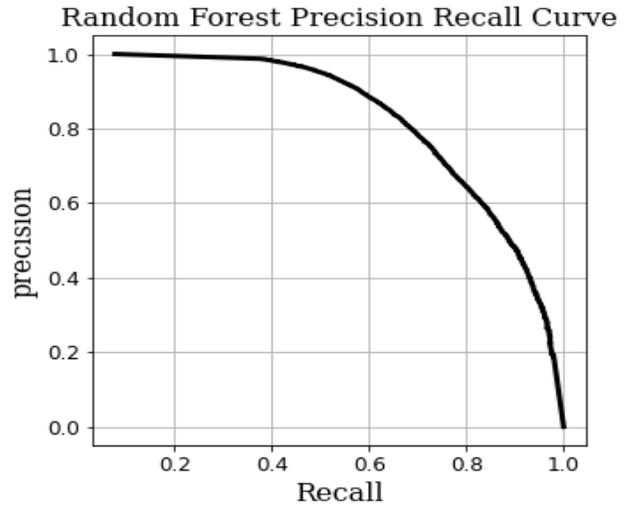

*Fig. 5 Precision-Recall-Curve for RF*

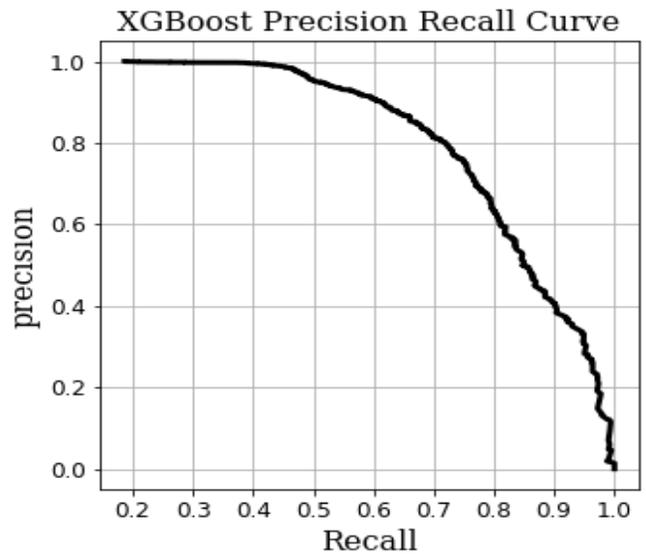

*Fig. 6 Precision-Recall-Curve for XGBoost*

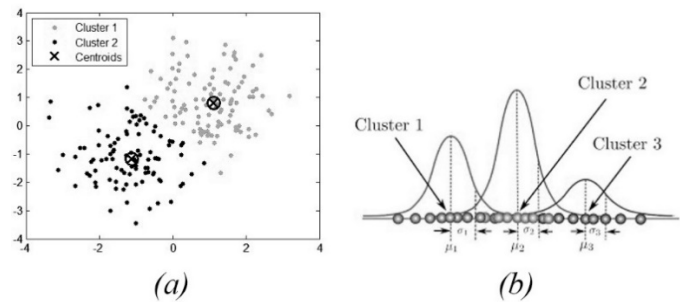

*Fig. 7 (a) K-means clustering which its operation is based on Euclidean distance while [18] (b) gaussian mixture model which works based on a density-based approach [19]*

In the propose SEMC-AD method, the embedding representation of alarm data are used as the input for PCA method. The dimensionality of data is reduced by applying the PCA and keeping the two significant principle components. By drawing the main components in 2D space, it is clear that the faults are densely packed in a certain corner of the new space. Fig. 8 and Fig. 9 illustrate it for train and test data. Here, using a proper clustering method, we found the cluster that has the highest density of faults. This cluster contains about 99% of the faults (Fig. 10). It is worth mentioning that the final clustering was performed by GMM with 5 clusters.

The results of SEMC-AD method show that the proposed approach is more sensitive to faults, since comparing with the baseline methods (RF and XGBoost), the value of precision is 63 % which increase slightly, while the value of recall is 99 %. The results demonstrate 17.8 % improvement in the performance of the proposed SEMC-AD method, in comparison to the baseline methods.

*TABLE 2*

*Models Performance*

| Methods | Class | Precision | Recall |
|---|---|---|---|
| Random Forest | 0 | 99 % | 78 % |
|  | 1 | 60 % | 86 % |
| XGBoost | 0 | 98 % | 83 % |
|  | 1 | 61 % | 81 % |
| SEMC-AD | 0 | 99 % | 76 % |
|  | 1 | 63 % | 99 % |

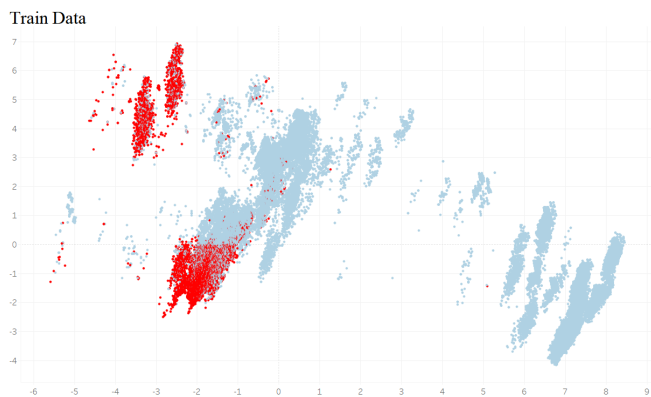

*Fig. 8 2D embedding representation of operator's behavior for 5 months of train data*

## IV. CONCLUSION

In this paper, a Supervised Embedding and Clustering Anomaly Detection (SEMC-AD) method is proposed, in which the principle components of the embedding vectors of alarm data are extracted and used for clustering the alarm data into two class of anomaly and normal. Considering the values of recall and precision metrics, the performance of the proposed SEMC_AD method is 18% better than the ordinary random forest and Gradient boosting method without embedding. 99% percent of anomalies are detected by SEMC-AD method while 86% and 81% of anomalies are detected by Random Forest and XGBoost, respectively.

Although it seems that in the labeled datasets, the supervised classification methods may have higher performance in anomaly detection, but the results demonstrate that in our dataset with high number of categorical features, the SEMC_AD method can classify the anomalies more efficiently.

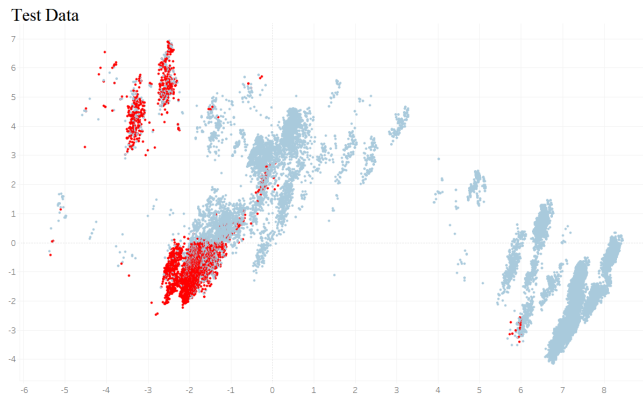

*Fig. 9 2D embedding representation of test data*

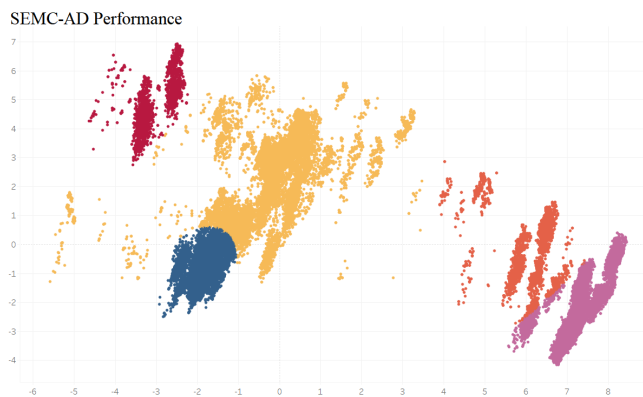

*Fig. 10 Clustering Performance illustration in 2D space*